\documentclass[conference]{IEEEtran}

\IEEEoverridecommandlockouts
\usepackage{cite}
\usepackage{amsmath,amssymb,amsfonts}
\usepackage{algpseudocode, algorithmicx}
\usepackage{algorithm}
\usepackage{graphicx}
\usepackage{float}
\usepackage{subfig}
\usepackage{textcomp}
\usepackage{xcolor}
\usepackage[export]{adjustbox}
\usepackage[a4paper, total={184mm,239mm}]{geometry}
\usepackage{colortbl}
\usepackage{makecell}
\usepackage{multirow}
\usepackage{soul}

\begin{document}
\title{Value-Driven Mixed-Precision Quantization for Patch-Based Inference on Microcontrollers
}

\author{
    \IEEEauthorblockN{Wei Tao$^{1}$, Shenglin He$^{1}$, Kai Lu$^{1}$, Xiaoyang Qu$^{2}$, Guokuan Li$^{1*}$\thanks{$^{*}$Guokuan Li (email: liguokuan@hust.edu.cn) and Jianzong Wang (email: jzwang@188.com) are the corresponding authors.}, Jiguang Wan$^{1}$, Jianzong Wang$^{2*}$, Jing Xiao$^{2}$}
    \IEEEauthorblockA{$^{1}$Wuhan National Laboratory for Optoelectronics, Huazhong University of Science and Technology, Wuhan, China}
    \IEEEauthorblockA{$^{2}$Ping An Technology (Shenzhen) Co., Ltd., Shenzhen, China}
    % \IEEEauthorblockA{\{D202281288, M202273717, kailu, jgwan,  liguokuan\}@hust.edu.cn, \{quxiaoyang343, wangjianzong347, xiaojing661\}@pingan.com.cn}
}

\maketitle

\begin{abstract}
Deploying neural networks on microcontroller units (MCUs) presents substantial challenges due to their constrained computation and memory resources. Previous researches have explored patch-based inference as a strategy to conserve memory without sacrificing model accuracy. However, this technique suffers from severe redundant computation overhead, leading to a substantial increase in execution latency. A feasible solution to address this issue is mixed-precision quantization, but it faces the challenges of accuracy degradation and a time-consuming search time. In this paper, we propose QuantMCU, a novel patch-based inference method that utilizes value-driven mixed-precision quantization to reduce redundant computation. We first utilize value-driven patch classification (VDPC) to maintain the model accuracy. VDPC classifies patches into two classes based on whether they contain outlier values. For patches containing outlier values, we apply 8-bit quantization to the feature maps on the dataflow branches that follow. In addition, for patches without outlier values, we utilize value-driven quantization search (VDQS) on the feature maps of their following dataflow branches to reduce search time. Specifically, VDQS introduces a novel quantization search metric that takes into account both computation and accuracy, and it employs entropy as an accuracy representation to avoid additional training. VDQS also adopts an iterative approach to determine the bitwidth of each feature map to further accelerate the search process. Experimental results on real-world MCU devices show that QuantMCU can reduce computation by 2.2x on average while maintaining comparable model accuracy compared to the state-of-the-art patch-based inference methods. 
\end{abstract}


\section{Introduction}
Neural networks have been widely used in various fields. Microcontroller units (MCUs) are a cost-effective and energy-efficient solution for neural network inference. Nonetheless, deploying neural network inference on MCUs poses substantial challenges owing to their restricted memory and computational resources. A typical MCU has a static random-access memory (SRAM) of less than 512 KB and a processing core with a frequency below 400MHz, which is inadequate for general neural networks.

Prior researches have employed various strategies to tackle this problem, including mixed-precision quantization \cite{kundu2022bmpq,wang2019haq, yao2021hawq, rusci2020memory, zhu2018adaptive}, efficient network architecture design \cite{sandler2018mobilenetv2, sun2022entropy, lin2021mcunetv2}, and dataflow scheduling \cite{lin2021mcunetv2, cipolletta2021dataflow, saha2020rnnpool, liberis2019neural, miao2022towards}. However, the mixed-precision quantization search process and the design of network architecture are time-consuming. Furthermore, mixed-precision quantization leads to accuracy loss. While dataflow scheduling does not result in accuracy loss, it is only suitable for a few neural networks.

\begin{figure*}
    \centering
    \subfloat[Patch-Based Inference Demostration]{
    \label{fig: patch-based demo}
        \begin{minipage}[b]{0.48\textwidth}
            \includegraphics[width=1\textwidth]{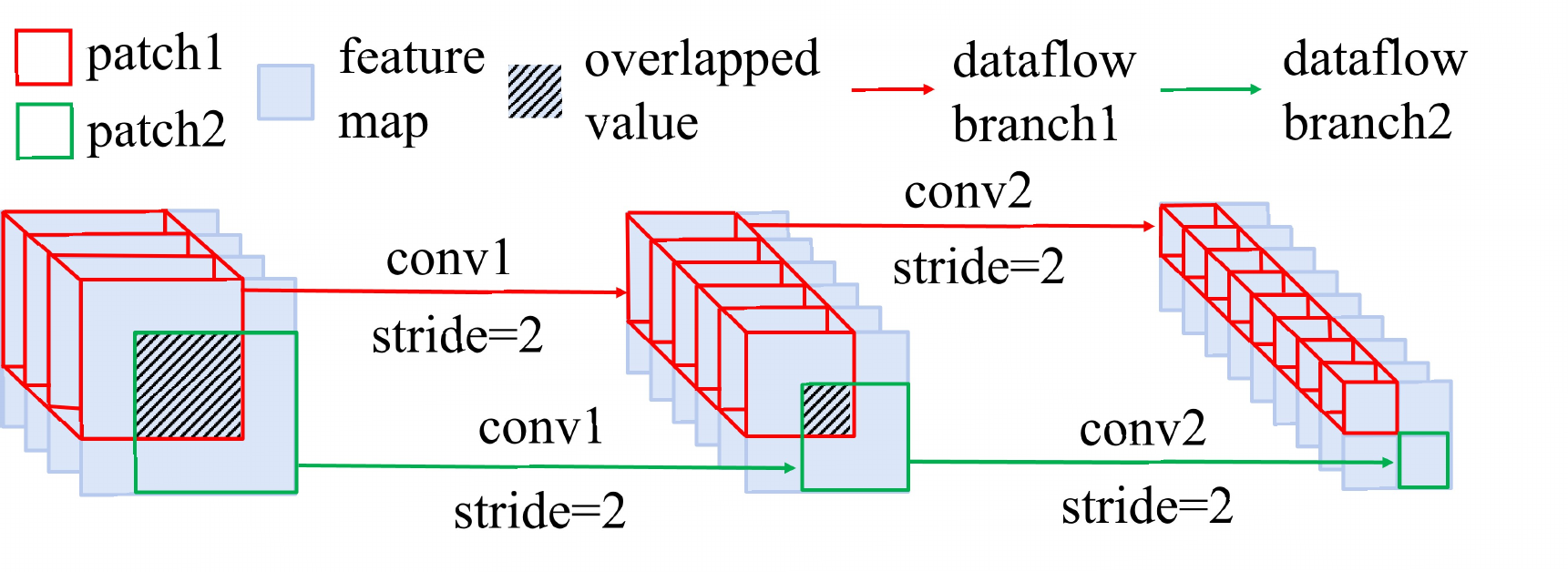}    
        \end{minipage}    
    }
    \subfloat[Inference Latency Comparison]{
    \label{fig: patch issue}
        \begin{minipage}[b]{0.48\textwidth}
            \includegraphics[width=1\textwidth]{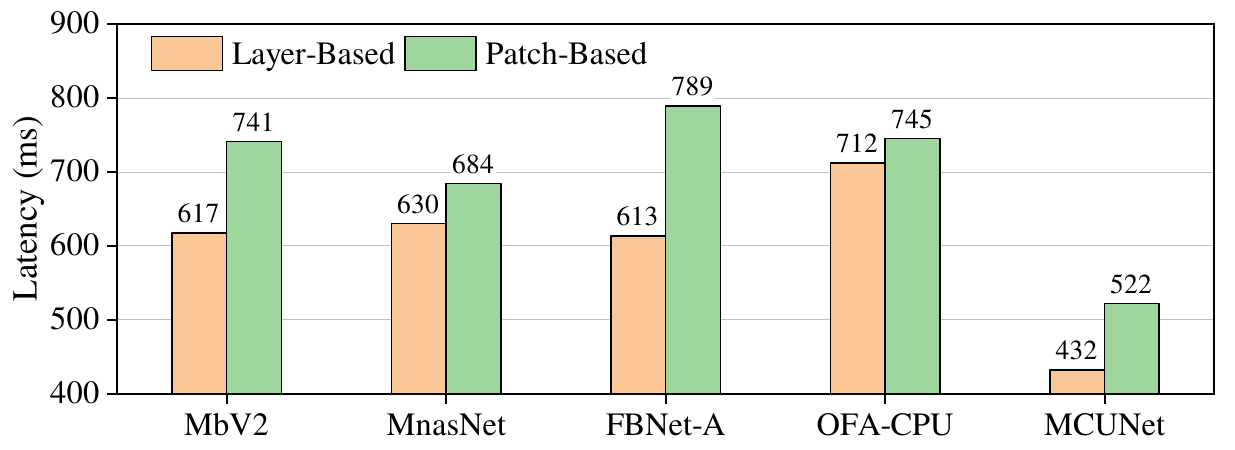}
        \end{minipage}    
    }
    \caption{\textbf{(a):} A simple demonstration of patch-based inference process (only two patches are drawn). \textbf{(b):} A comparison experiment result of the inference latency of patch-based and layer-based inference.}
    \label{fig: background}
\end{figure*}

Recently, a new dataflow scheduling method known as patch-based inference \cite{lin2021mcunetv2,cipolletta2021dataflow, saha2020rnnpool} has emerged. Figure \ref{fig: patch-based demo} demonstrates the patch-based inference computation process. Patch-based inference splits the input feature map 
% (Feature map is also called activation value. In the rest of this paper, we will use these two expressions interchangeably)
into multiple patches, causing the neural network's dataflow to be divided into branches, each following a patch. Neural network inference involves sequential execution of dataflow branches. Each branch has a lower peak memory use compared to the overall dataflow. Patch-based inference successfully reduces the neural network's memory footprint. 
Patch-based inference is a common dataflow scheduling technique that may be used in nearly all neural networks. However, patch-based inference presents a severe issue of redundant computation. As is illustrated in Figure \ref{fig: patch-based demo}, there exist overlapped values in the dataflow branches. Overlapped values are computed twice, leading to increased computation and inference latency. Layer-based inference is the traditional inference method, where the neural network is executed layer-by-layer. We compared the latency of layer-based and patch-based inference on different models. Figure \ref{fig: patch issue} demonstrates that patch-based inference leads to an 8$\%$-17$\%$ increase in inference latency. Previous patch-based inference approaches used heuristic methods \cite{cipolletta2021dataflow, lin2021mcunetv2} to address this issue, resulting in suboptimal results. Mixed-precision quantization can reduce redundant computation due to the fact that values with lower bitwidth require less computation. However, it can lead to accuracy loss and time-consuming search processes.

To address these issues, we propose QuantMCU, a novel patch-based inference approach that uses value-driven mixed-precision quantization to reduce redundant computation. First, QuantMCU performs value-driven patch classification (VDPC) to ensure model accuracy. VDPC divides patches into two categories: outliers and non-outliers.  We apply 8-bit quantization for outlier class patches and feature maps on the following dataflow branches. QuantMCU uses a value-driven quantization search (VDQS) strategy on non-outlier class patches and feature maps in the following dataflow branches to reduce search time. VDQS introduces a new search measure that incorporates both accuracy and computation. Model accuracy is represented by activation value entropy, eliminating the need to train the model at each search step. Furthermore, VDQS uses a lightweight iterative technique to determine the bitwidth for each feature map. By applying VDPC and VDQS, we can effectively reduce the redundant computation of the model in patch-based inference and further reduce its peak memory usage, with comparable model accuracy.

The main contributions of our work are summarized below.
\begin{itemize}

\item We propose value-driven patch classification (VDPC) for the feature map patches in patch-based inference (\textbf{Section \ref{subsec:classify}}) to maintain model accuracy during our quantization.

\item We apply value-driven quantization search (VDQS) on the outlier class patches and the feature maps on the dataflow branches that follow (\textbf{Section \ref{subsec:search}}). VDQS can effectively shorten the search time of mixed-precision quantization.

\item We implement QuantMCU on real-world MCU devices with different resource constraints. We test QuantMCU on various neural networks and two standard datasets. Extensive experiments show that QuantMCU can, on average, reduce the BitOPs and inference latency of state-of-the-art patch-based inference methods by 2.2x and 1.5x, respectively.

\end{itemize}

\section{Related Works} 
\label{sec:related works}
\textbf{Neural network inference on MCUs.}
The application of neural network inference on MCUs has shown rapid growth in recent years. Many inference frameworks have been designed, including Tensorflow Lite Micro \cite{abadi2016tensorflow}, CMSIS-NN \cite{lai2018cmsis}, CMix-NN \cite{capotondi2020cmix}, TinyEngine \cite{lin2021mcunetv2}, etc. However, none of these frameworks comprehensively considers the optimization for memory usage and computation. Due to these frameworks, neural networks are either unable to deploy because of excessive memory utilization, or if they do, the inference latency will be intolerable. 

\textbf{Dataflow scheduling.}
The memory footprint of a neural network during inference is closely related to its dataflow. Recent attempts have been made to reduce the peak memory usage of neural networks by scheduling their dataflow. Some researchers \cite{liberis2019neural} try to reorder the operator inferences. They search for an optimal topology of the dataflow graph manually, which represents the smallest peak memory usage. Miao et al. \cite{miao2022towards}  propose to swap data between SRAM and external storage. However, the above dataflow scheduling methods cannot be applied to all the neural networks. Other works \cite{lin2021mcunetv2,cipolletta2021dataflow,saha2020rnnpool} focus on patch-based inference. For example, Cipolletta et al. \cite{cipolletta2021dataflow} design a restructuring algorithm to find the optimal patch split layer and dataflow branch length. Saha et al. \cite{saha2020rnnpool} design a new pooling operator to compute partial feature maps across multiple layers. Nonetheless, there exists a severe redundant computation issue in patch-based inference methods, and their heuristic solutions cannot address it well. In this work, we integrate patch-based inference with value-driven mixed-precision quantization, resulting in a significant reduction in the model's redundant computation and further decreasing its peak memory usage.

\textbf{Mixed-precision quantization.}
Quantization, which has long been a popular method of compressing models, is the process of converting floating point 32-bit (FP32) type data in neural networks into lower-bit type \cite{qu2020quantization}. Mixed-precision quantization methods \cite{kundu2022bmpq,wang2019haq, yao2021hawq, rusci2020memory, sun2022entropy,zhu2018adaptive} aim to assign different bitwidth (not higher than 8-bit) for the data of different layers based on the observation that different layers in a neural network have different sensitivity to accuracy. For example, Wang et al. use reinforcement learning (RL) \cite{wang2019haq} to search for the quantization configuration, which is effective but requires a lot of time and computation resources. The goal of HAWQ-V3 \cite{yao2021hawq} is to allocate bandwidth for each layer based on specific and easily derived metrics, e.g. Hessian spectrum. However, this method fails to consider the change of sensitivity when the values are being quantized or updated in the quantization-aware training process. While Rusci et al. \cite{rusci2020memory} have proposed efficient per-channel quantization, their quantization does not take model accuracy optimization into account. Some researchers \cite{sun2022entropy, zhu2018adaptive} have tried to use entropy as the agent of model accuracy to achieve better gradient approximation and lower computation cost. Nonetheless, they usually rely on complicated search techniques such as neural architecture search (NAS). In summary, traditional mixed-precision quantization methods suffer from accuracy loss and usually require time-consuming search processes. In this work, We propose a value-driven mixed-precision quantization approach that utilizes VDPC to maintain excellent model accuracy and employs VDQS to significantly shorten search time.

\section{Proposed method}
\label{sec:method}

In this section, the value-driven patch classification and the value-driven quantization search in QuantMCU will be described in detail.
We will first introduce how we classify patches according to the outlier value. Then, the design of a new quantization search metric and a lightweight iterative search algorithm will be described. 

\subsection{Value-driven patch classification}
\label{subsec:classify}

\begin{figure}
    \centering
    \subfloat[]{
    \label{fig: distribution}
        \begin{minipage}[b]{0.22\textwidth}
            \includegraphics[width=1\textwidth]{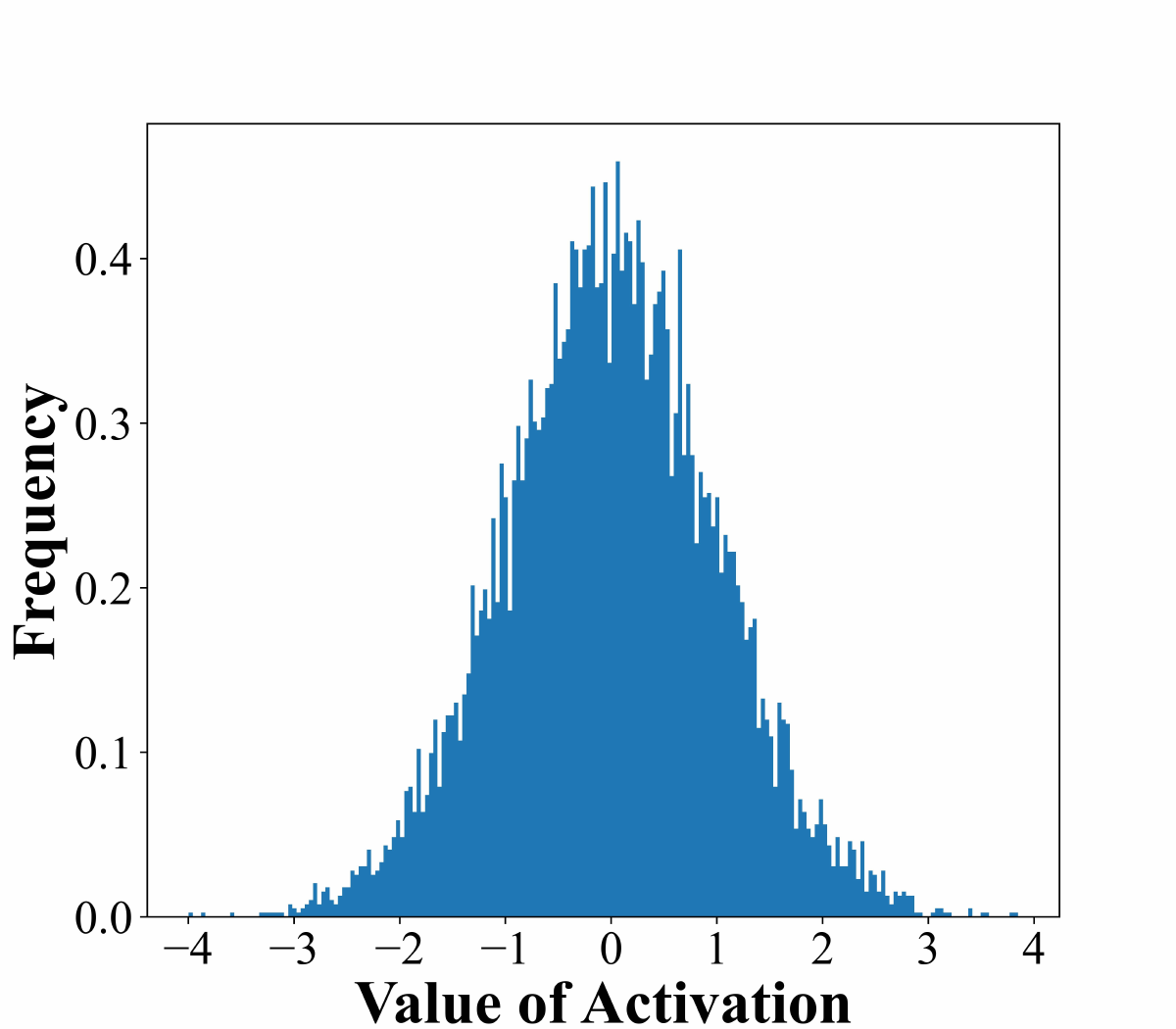}
        \end{minipage}
    }
    \subfloat[]{
    \label{fig: outlier value}
        \begin{minipage}[b]{0.22\textwidth}
            \includegraphics[width=1\textwidth]{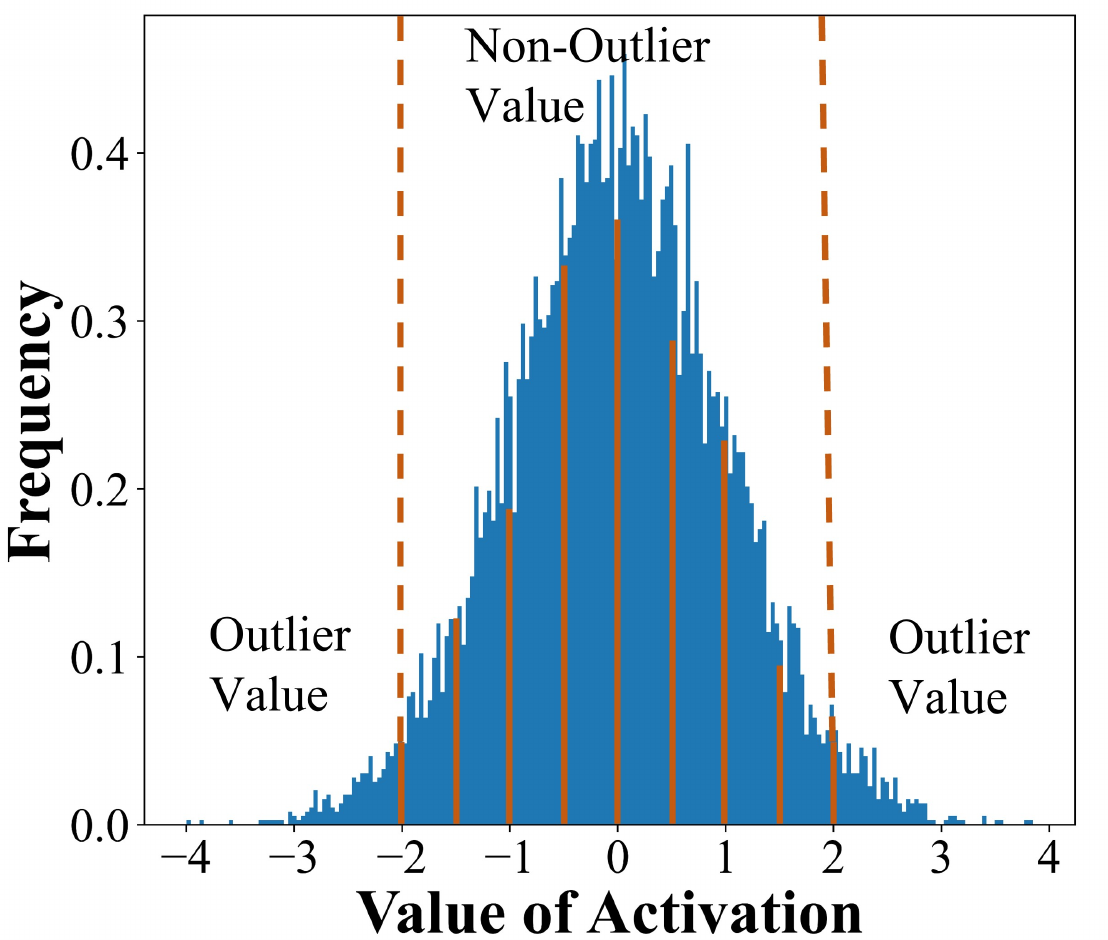}
        \end{minipage}
    }
    \caption{\textbf{(a):} The distribution of the output activation value of the first layer in ResNet18. \textbf{(b):} The separation of outlier value and non-outlier value.}
    \label{fig: outlier threshold}
\end{figure}

In general, the activation value distribution of neural networks exhibits a bell-shaped or Gaussian-like pattern, as is illustrated in Figure \ref{fig: distribution}. One notable characteristic of such distribution is that while a small portion of values deviates significantly from 0 and often plays a crucial role, the majority of values cluster around 0 and have minimal impact on model accuracy.

\begin{figure}
    \centering
    \includegraphics[width=0.48\textwidth]{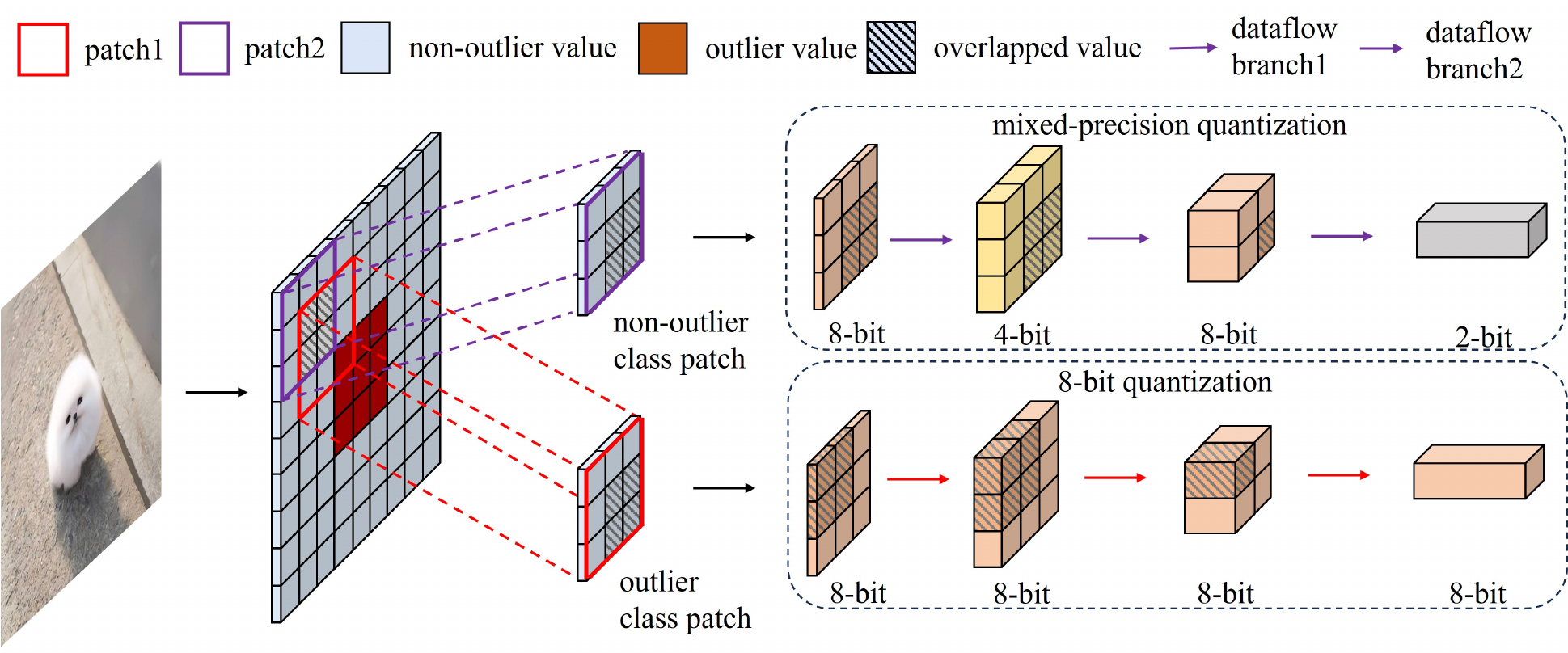}
    \caption{A demonstration of VDPC. Patch1 is classified as an outlier class patch since it contains an outlier value at its bottom right corner. Patch2 is classified as a non-outlier class patch since it does not contain any outlier value. For patch1 and dataflow branch1, we apply 8-bit quantization, while for patch2 and dataflow branch2, we apply mixed-precision quantization.}
    \label{fig: seperate type}
\end{figure}

Based on the idea presented in \cite{guo2022ant}, we designate the value around 0 as the non-outlier value, and the value far away from 0 as the outlier value, as is illustrated in Figure \ref{fig: outlier value}. Since activation values are composed of outlier values and non-outlier values, each split patch may consist entirely of outlier values, entirely non-outlier values, or maybe both. As a result, the patches are divided into \textbf{outlier class} and \textbf{non-outlier class} using VDPC. The patches with outlier values are known as the outlier class patches. Applying mixed-precision quantization to the outlier class patches will significantly affect model accuracy because outlier values play a major role in accuracy. Since 8-bit quantization can preserve more model information than mixed-precision quantization, we only apply it to outlier class patches. Feature maps on the dataflow branches that follow should also only use 8-bit quantization. The VDPC method will avoid a significant accuracy decrease in the quantization.
On the other hand, the non-outlier class patches are those patches full of non-outlier values. Since non-outlier values are less important in terms of accuracy, we apply mixed-precision quantization to these patches and the feature maps on the dataflow branches following them. Figure \ref{fig: seperate type} shows a VDPC demonstration.

Assuming that the activation value distribution follows a Gaussian distribution, we compute its Probability Density Function (PDF) to find the best outlier/non-outlier separation. For each activation value $x$, there is:

\begin{equation}
    \label{eqn:gaussian pdf}
    F(x) =
    \left\{
    \begin{array}{lr}
        0, & {\frac{1}{\sqrt{2\pi \sigma^2}}e^{-\frac{(x-\mu)^2}{2\sigma^2}} \leq \phi}\\
        1, & {\frac{1}{\sqrt{2\pi \sigma^2}}e^{-\frac{(x-\mu)^2}{2\sigma^2}} > \phi}\\
    \end{array}
    \right.
\end{equation}

where $F(x)=1$ indicates an outlier value and $F(x)=0$ indicates a non-outlier value, $\mu$ is the mean, $\sigma$ is the variation, and $\phi$ is a predefined threshold. $\phi$ should be properly set. An excessively large $\phi$ will eliminate some important information conveyed by the outlier value and result in a sharp decline in accuracy, while an extremely small $\phi$ can not fully reduce the redundant computation.
We test different configurations of $\phi$ in the practical experiment  (\textbf{Section \ref{subsec: analysis}}).

\subsection{Value-driven quantization search}
\label{subsec:search}

\textbf{Quantization score.}
It should be noted that the accuracy and computation of the model will change each time quantization is applied to a feature map. To accurately measure the impact, VDQS specifies a new search metric called quantization score. 

First, we use Bit Operations (BitOPs) as the representation of computation. If the $i_{th}$ feature map is quantized to $b$ bitwidth, we define the impact on the model computation as follows:
\begin{equation}
    \Phi(i, b) = \frac{\Delta{B(i, b)}}{B}
\end{equation}
where $\Delta{B(i, b)}$ is the BitOPs reduction of the $i_{th}$ feature map after $b$-bit quantization, and $B$ is the sum of the BitOPs of all the feature maps in the full-precision model. 

Subsequently, we employ activation value entropy as the representation of accuracy so as to avoid training the model at each step of the quantization. Our insight is that a quantized feature map with higher entropy can preserve more representative capabilities of the original model \cite{zhu2018adaptive}. Furthermore, the entropy of the last output feature map determines the expressiveness of the system \cite{sun2022entropy}. In order to calculate the entropy of the $i_{th}$ feature map with bitwidth $b$, we use the empirical distribution to approximate the actual distribution of the activation values. In other words, we divide the activation value range uniformly into $k$ bins, where $k$ is a predefined hyperparameter. We count the number of activation values that fall into the $j_{th}$ bin as $x_j$. Then we assume that each activation value in the $j_{th}$ bin has the probability:
\begin{equation}
    \hat{p}_j = \frac{x_j}{n_i}
\end{equation}
where $n_i$ is the total number of activation values in the $i_{th}$ feature map. The entropy of the $i_{th}$ feature map with bitwidth $b$ is calculated as follows:
\begin{equation}
    H(i,b) = -\sum_j{\hat{p}_jlog(\hat{p}_j)}
\end{equation}
We define the impact on the model accuracy as follows:
\begin{equation}
    \Omega(i, b) = \frac{\Delta{H(i,b)}}{H(N,b_{last})}
\end{equation}
where $\Delta{H(i,b)}$ is the entropy reduction of the $i_{th}$ feature map after $b$ bitwidth quantization, and $H(N,b_{last})$ is the entropy of the last output feature map with bitwidth $b_{last}$.

Finally, we define quantization score as the weighted sum of $\Phi(i,b)$ and $\Omega(i,b)$, which is calculated as follows:
\begin{equation}
    S(i,b) = -\lambda\Omega(i,b) + (1-\lambda)\Phi(i,b)
    \label{eqn:quantization score}
\end{equation}
where $\lambda$ is the weight parameter that balances the importance of computation and accuracy. 
$\lambda$ is chosen based on how current applications appreciate the neural network's computation and accuracy. $S(i,b)$ represents the benefits of $b$ bitwidth quantization to the $i_{th}$ feature map.

\textbf{Lightweight iterative search process.}
VDQS utilizes a lightweight iterative search algorithm based on quantization score to determine the bitwidth of each feature map. The detailed algorithm is shown in Algorithm \ref{alg1}. For every feature map, there are $m$ kinds of candidate bitwidths(In a real implementation, due to the constraint of the software library, the feature map is only able to be quantized to 8, 4, and 2 bits. Therefore $m$ is set to be 3 in practice.). First, we calculate the quantization score for every feature map with every possible bitwidth. Then we select each feature map with the highest quantization score as the initial bitwidth. It should be noted that every two adjacent feature maps should satisfy the following constraint:
\begin{equation}
    {\rm Mem} (i,b_i) + {\rm Mem} (i+1, b_{i+1}) \leq M  \quad i=0,1,...,N-1 
    \label{eqn:restrict}
\end{equation}
where Mem($i,b_i$) denotes the memory usage of the $i_{th}$ feature map with bitwidth $b_i$, and $M$ denotes the memory constraint of the MCU.
Finally, we iterate the dataflow branch until Equation \ref{eqn:restrict} is satisfied. During the two iterations, we consider the pair that consists of two adjacent feature maps. We adjust the latter feature map in the first iteration and the former feature map in the second iteration. To make the necessary change, the bitwidth of the feature map with the suboptimal quantization score in comparison to the present one is set.

\begin{algorithm}
\caption{Bitwidth determination}
\label{alg1}
\begin{algorithmic}[1]
\Require A dataflow branch of $N$ layers, a memory constraint $M$, available bitwidth kinds $m$, candidate bitwidths sets for each feature map: $\{s_1^i, s_2^i,...,s_m^i\}_{i=0}^N$
\Ensure The bitwidth for each feature map: $b_0, b_1, ..., b_N$
\For{$i = 0$ to $N$}
\For{$j = 1$ to $m$}
\State calculate $S(i,s_j^i)$ according to Equation \ref{eqn:quantization score}
\EndFor
\State sort ${s_1^i, s_2^i,...s_m^i}$ according to the descending order of quantization score and derive a new set ${t_1^i, t_2^i,...,t_m^i}$
\State $b_i \leftarrow t_1^i$
\EndFor
\While{Equation \ref{eqn:restrict} is not True}
    \State \Call{Traverse}{0, N-1, 1}
    \State \Call{Traverse}{1, N, -1}
\EndWhile

\Function{Traverse}{$a, b, r$}
    \For{$i = a$ to $b$}
        % \State $j \leftarrow$ the index of $b_i$ in $t_1^i, t_2^i,..., t_m^i$
        \State Denote the index of $b_{i+r}$ in $t_1^{i+r}, t_2^{i+r},..., t_m^{i+r}$ as $k$
        \While{\Call{NeedChange}{i, r, k}}
            % \If{$k < m$ and Mem($i, b_i$) $\leq$ Mem($i+r, b_{i+r}$)} 
            \State $b_{i+r} \leftarrow t_{k+1}^{i+r}$
            % \EndIf
        \EndWhile
    \EndFor
\EndFunction

\Function{NeedChange}{$i, r, k$}
    \If{Mem($i, b_i$) $+$ Mem($i+1, b_{i+1}$) $> M$}
        \If{$k < m$ and Mem($i, b_i$) $\leq$ Mem($i+r, b_{i+r}$)}
            \State \Return True
        \EndIf
    \EndIf
\State \Return False
\EndFunction

\end{algorithmic}

\end{algorithm}

\section{Evaluation Results}
\label{sec:evaluation}
In this section, we first present the experiment results of QuantMCU with layer-based and patch-based inference methods on image classification tasks and object detection tasks on two MCU platforms. Then we conduct ablation studies to validate the impact of VDPC and VDQS. Finally, we analyze the influence of hyperparameters and visualize the quantization results of QuantMCU.

\subsection{Experiment setup}
\label{subsec:experiment setup}

\textbf{Datasets.} We conduct the experiments on two classic neural network applications: image classification and object detection to evaluate QuantMCU in various fields. We use two standard benchmarks in this work: ImageNet \cite{deng2009imagenet} for the image classification task and Pascal VOC \cite{everingham2010pascal} for the object detection task. All the images are resized to a resolution of 224*224.

\textbf{Implementation.} We have tested QuantMCU on two different real-world MCU devices: Arduino Nano 33 BLE Sense
(ARM Cortex-M4, 256kB SRAM/1MB Flash) and STM32H743 (ARM Cortex-M7, 512KB SRAM/2MB Flash). We use Tensorflow Lite \cite{abadi2016tensorflow} to execute 8-bit quantization and the CMix-NN library \cite{capotondi2020cmix} for sub-byte quantization on MCUs.

\begin{table*}[t]
    \centering
    \caption{The comparison of QuantMCU with state-of-the-art patch-based inference methods and layer-based inference methods on MobileNetV2 network on different datasets and different platforms. The width multiplier and resolution of the model are adjusted to fit MCU memory.}
    \begin{tabular}{cccccccc}
    \hline
    \textbf{Platform} & \textbf{Dataset} & \textbf{Metric} & \textbf{Layer-Based} & \textbf{MCUNetV2 \cite{lin2021mcunetv2}} & \textbf{Cipolletta et al. \cite{cipolletta2021dataflow}} & \textbf{RNNPool \cite{saha2020rnnpool}} & \textbf{QuantMCU}  \\
    \hline
    \multirow{6}{*}{\textbf{\makecell{Arduino Nano \\ 33 BLE Sense \\ (256KB SRAM, \\ 1MB Flash)}}} & \multirow{3}{*}{\makecell{ImageNet \\ \cite{deng2009imagenet}}} 
     & Peak Memory (KB) & 244 & 196 & 122 & 226 & \textbf{78} \\
    & & BitOPs (M) & 1536 & 1690 & 1721 & 1582 & \textbf{719} \\
    & & inference Lat. (ms) & 617 & 741 & 784 & 640 & \textbf{486} \\
     \cline{2-8}
   &  \multirow{3}{*}{\makecell{PascalVOC \\ \cite{everingham2010pascal}}} 
    & Peak Memory (KB) & 252 & 207 & 146 & 242 & \textbf{99} \\
    & & BitOPs (M) & 2176 & 2459 & 2524 & 2389 & \textbf{1171} \\
    & & inference Lat. (ms) & 656 & 793 & 848 & 717 & \textbf{502} \\
     \hline
    \multirow{6}{*}{\textbf{\makecell{STM32H743 \\ (512KB SRAM, \\ 2MB Flash)}}} & \multirow{3}{*}{\makecell{ImageNet \\ \cite{deng2009imagenet}}}  & Peak Memory (KB) & 505 & 465 & 380 & 477 & \textbf{298} \\
    & & BitOPs (M) & 4057 & 4283 & 4405 & 4124 & \textbf{1987} \\
    & & inference Lat. (ms) & 1684 & 1799 & 1945 & 1736 & \textbf{1208}\\
     \cline{2-8}
   &  \multirow{3}{*}{\makecell{PascalVOC \\ \cite{everingham2010pascal}}}  & Peak Memory (KB) & 509 & 438 & 382 & 477 & \textbf{303} \\
    & & BitOPs (M) & 5842 & 6162 & 6347 & 5938 & \textbf{2933} \\
    & & inference Lat. (ms) & 1792 & 1912 & 2089 & 1836 & \textbf{1323} \\
     \hline
    \end{tabular}
    \label{tab: performance comparison}
    \vspace{-2em}
\end{table*}

\begin{figure*}
    \centering
    \subfloat[Top-1 Accuracy on ImageNet]{
    \label{fig: accuracy1}
        \begin{minipage}[b]{0.48\textwidth}
            \includegraphics[width=1\textwidth]{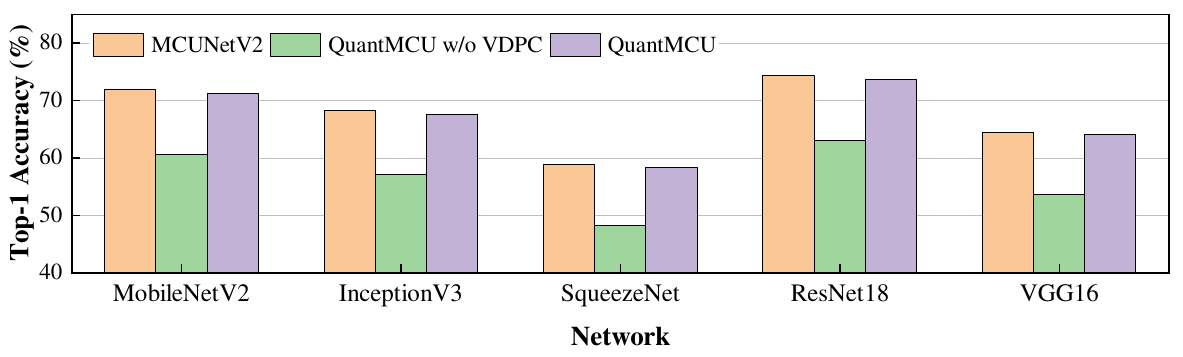}    
        \end{minipage}    
    }
    \subfloat[mAP on Pascal VOC]{
    \label{fig: accuracy2}
        \begin{minipage}[b]{0.48\textwidth}
            \includegraphics[width=1\textwidth]{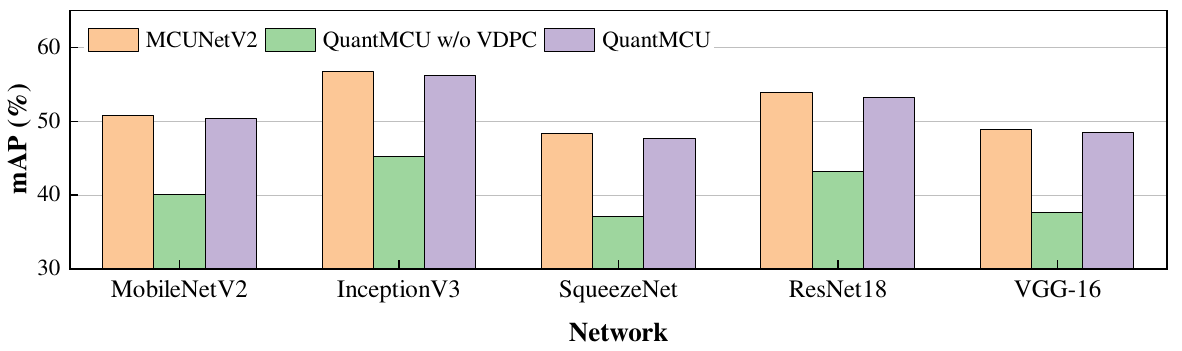}
        \end{minipage}    
    }
    \caption{The accuracy comparison of QuantMCU with patch-based inference on different networks on two different datasets.}
    \label{fig: accuracy comparison}
    \vspace{-1em}
\end{figure*}

\textbf{Metrics.} We use BitOPs to evaluate the model computation. For the image classification task, we employ the Top-1 correct rate as the accuracy measure. For the object detection task, mAP (mean Average Precision) is used to evaluate accuracy. We also measure the inference latency of neural networks using QuantMCU and patch-based inference approaches on MCU devices.

\subsection{Comparison with state-of-the-art patch-based inference methods}
We evaluate the performance of QuantMCU against three state-of-the-art patch-based inference methods and a layer-based inference method. The results are shown in Table \ref{tab: performance comparison}. The table shows that QuantMCU may further reduce the peak memory usage of patch-based inference methods by 1.26x-2.90x. Furthermore, all the patch-based inference methods have higher BitOPs and inference latency than the layer-based inference method. However, QuantMCU can reduce the BitOPs and inference latency of patch-based inference methods, even smaller than those of the layer-based inference method. QuantMCU decreases the BitOPs and the inference latency by 2.2x and 1.5x respectively on average compared to the patch-based inference methods. This is mainly due to our combination of quantization techniques.

\begin{table}[b]
    \begin{center}
    \caption{Comparison of different quantization methods on MobileNetV2 network on ImageNet dataset. ``W/A-Bits" means the bitwidth of weight/activation value. ``MP" means mixed-precision. ``Time" means the time cost of quantization process.}
    \begin{tabular}{ccccccc}
    \hline
       \textbf{Method} & \textbf{W/A-Bits} & \textbf{Top-1} & \textbf{BitOPs} & \textbf{Memory} & \textbf{Time}  \\
       \hline
        Baseline                    & 8/8         & 71.9\% & 19.2G  & 1372kB & -  \\ \hline
        Pact \cite{choi2018pact} & 4/4        & 61.4\% & 7.42G   & 692kB & 45min  \\
        Rusci et al. \cite{rusci2020memory}    & MP/MP         & 61.8\% & 7.42G   & 690kB & 33min   \\
        HAQ \cite{wang2019haq}      & MP/MP     & 68.5\% & 42.8G   & 950kB & 90min  \\
        HAWQ-V3 \cite{yao2021hawq} & MP/MP & 63.4\% & 13.6G  & 787kB & 30min   \\ 
        QuantMCU                         & 8/MP & \textbf{69.2\%} &  10.9G & \textbf{523kB} & \textbf{0.5min}   \\
        \hline
    \end{tabular}
    \label{tab: quantization comparison}
    \end{center}
    \vspace{-2em}
\end{table}

\subsection{Ablation study}
\noindent\textbf{Impact of VDPC.} We compare the accuracy of MCUNetV2 \cite{lin2021mcunetv2} (a classic patch-based inference method), ``QuantMCU without VDPC", and QuantMCU on two datasets. In the ``QuantMCU without VDPC" group we apply VDQS to all the patches. The results are shown in Figure \ref{fig: accuracy comparison}. ``QuantMCU without VDPC" experiences 10\%-15\% accuracy loss compared with MCUNetV2. In contrast, QuantMCU achieves less than 1\% accuracy loss on both datasets, demonstrating how VDPC preserves the model accuracy. 

\noindent\textbf{Impact of VDQS.} We compare VDQS with some state-of-the-art quantization methods, including uniform-precision and mixed-precision. The results are shown in Table \ref{tab: quantization comparison}. We can observe that VDQS reduces the peak memory usage of other quantization methods by 1.32x-2.62x. Besides, VDQS  increases the accuracy by 0.7\%-7.8\% compared with other quantization methods. In addition, VDQS can finish the quantization process in 0.5 minutes, which is drastically faster than other methods. This is because VDQS uses entropy as the representation of accuracy to avoid extra training and applies an iterative search process instead of relying on time-consuming RL or NAS.

\subsection{Analysis}
\label{subsec: analysis}

\begin{figure}
    \centering
    \includegraphics[width=0.48\textwidth]{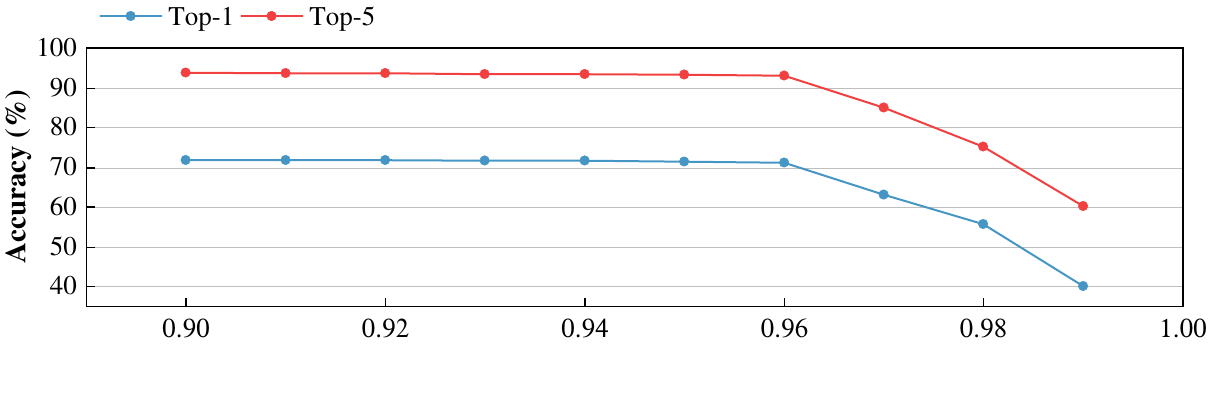}
    \caption{Top-1 and Top-5 accuracy of QuantMCU under different $\phi$ values on MobileNetV2 network on the ImageNet dataset.}
    \label{fig: parameter setting}
    \vspace{-1.2em}
\end{figure}

We measure the Top-1 and Top-5 accuracy on ImageNet with QuantMCU under different $\phi$. As is illustrated in Figure \ref{fig: parameter setting}, the Top-1 and Top-5 accuracy stays stable when $\phi$ is smaller than 0.96. However, when $\phi$ exceeds 0.96, the accuracy decreases rapidly. Therefore, we choose 0.96 as the optimal value of $\phi$. In addition, we test the impact of hyperparameter $\lambda$ on QuantMCU. As is shown in Table \ref{tab: lambda}, when $\lambda$ increases, the BitOPs rise along with the Top-1 Accuracy. We choose 0.6 as the value of $\lambda$ to achieve the best comprehensive benefit.

\begin{figure}[t]
    \centering
        \label{fig: bitwidth visualize}
        \includegraphics[width=0.48\textwidth]{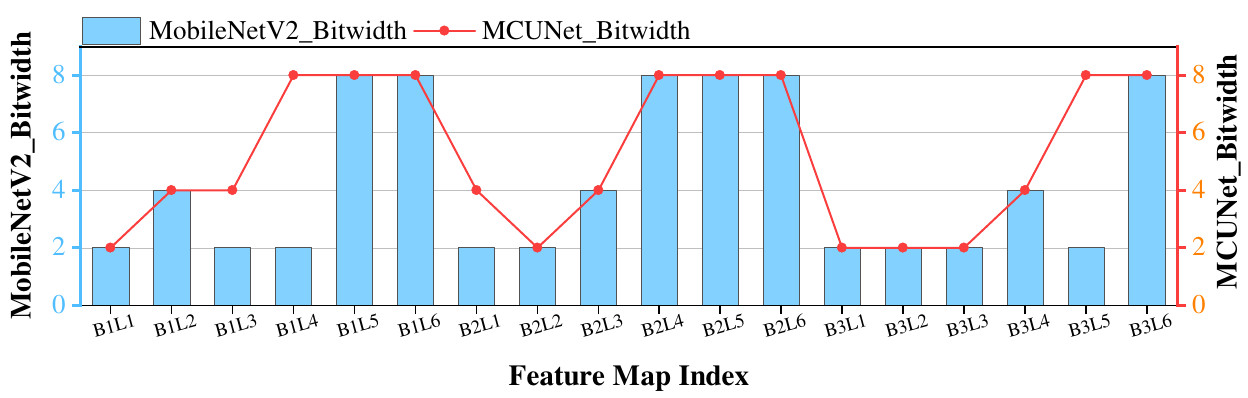} 
    \caption{The visualization of bitwidth assignment after quantization for MobileNetV2 and MCUNet network. The "BxLy" means the $y_{th}$ feature map on the $x_{th}$ branch.}
    \label{fig: analysis}
\end{figure}

We visualize the bitwidth assignment results after quantization in QuantMCU for MobileNetV2 and MCUNet in Figure \ref{fig: analysis}. From the figure, we can see that more than half of the feature maps are assigned sub-byte precision. The bitwidths of the layers at the end of a branch are mainly 8-bit, while the feature maps at the start of a branch are usually assigned low bitwidths. This is because the first few feature maps usually have a large size and need to be quantized to lower bitwidths in order to reduce computation, whereas the last few feature maps typically contribute significantly to model accuracy and require quantization to higher bitwidths to maintain accuracy.

\section{Conclusion}
\label{sec:conclusion}
We propose a novel value-driven mixed-precision quantization called QuantMCU for patch-based inference on MCUs. We utilize VDPC to maintain accuracy in the quantization, which classifies patches according to whether they contain outlier values. Besides, we employ VDQS to decrease the quantization search time. VDQS defines a new search measure based on activation value entropy and BitOPs to avoid additional training, and it uses a lightweight iterative search method to speed up the search process.  Experimental results on real-world MCU devices prove that QuantMCU can reduce the computation of previous patch-based inference methods by 2.2x on average.

\begin{table}[t]
    \centering
    \caption{The impact of different values of $\lambda$ on QuantMCU.}
    \begin{tabular}{ccccccccccc}
    \hline
    \textbf{$\lambda$}  & 0.2 & 0.3 & 0.4 & 0.5 & 0.6 & 0.7 & 0.8  \\
    \hline
    \textbf{Top-1 Acc. (\%)} & 65.6 & 67.1 & 67.9 & 68.7 & 69.2 & 70.1 & 71.2 \\
    \textbf{BitOPs (G)} & 7.6 & 8.4 & 9.2 & 10.1 & 10.9 & 14.3 & 18.7 \\
    \hline
    \end{tabular}
    \label{tab: lambda}
\end{table}

\section{Acknowledgements}
This work was sponsored by the Key Research and Development Program of Guangdong Province under Grant No. 2021B0101400003, the National Natural Science Foundation of China under Grant No.62072196, and the Creative Research Group Project of NSFC No.61821003. The corresponding authors are Guokuan Li from Huazhong University of Science and Technology (liguokuan@hust.edu.cn) and Jianzong Wang from Ping An Technology (Shenzhen) Co., Ltd. (jzwang@188.com).

\bibliographystyle{IEEEtran}
\bibliography{IEEEabrv,sample-base}

\end{document}